\documentclass[runningheads]{llncs}

\usepackage{eccv}

\usepackage{eccvabbrv}

\usepackage{graphicx}
\usepackage{booktabs}

\usepackage[accsupp]{axessibility}  

\usepackage[pagebackref,breaklinks,colorlinks,citecolor=eccvblue]{hyperref}

\usepackage{orcidlink}

\usepackage{todonotes}
\usepackage{comment}
\usepackage{amsmath,amssymb}
\usepackage{color}
\usepackage{url}
\usepackage{hyperref}
\usepackage{subcaption}

\usepackage{multirow}
\usepackage{xcolor}
\usepackage{siunitx}
\usepackage{etoolbox}
\robustify\bfseries
\usepackage{etoolbox}
\robustify\itshape
\usepackage{longtable}
\usepackage{diagbox}

\usepackage{multirow}

\begin{document}

\title{AnomalousPatchCore: Exploring the Use of Anomalous Samples in Industrial Anomaly Detection}

\titlerunning{AnomalousPatchCore}

\author{Mykhailo Koshil\inst{1}\orcidlink{0009-0004-1652-0446},
Tilman Wegener\inst{2} \orcidlink{0000-0001-8502-1238},
Detlef Mentrup\inst{2}, \\ Simone Frintrop\inst{3}\orcidlink{0000-0002-9475-3593}, Christian Wilms\inst{3}\orcidlink{0009-0003-2490-7029}}


\authorrunning{M.~Koshil et al.}

\institute{Department of Computer Science, University of Tübingen, Germany  \and
Basler AG, An der Strusbek 60-62, 22926 Ahrensburg, Germany \and
Computer Vision Group, University of Hamburg, Germany\\
\email{mykhailo.koshil@uni-tuebingen.de}
}

\maketitle

\begin{abstract}
Visual inspection, or industrial anomaly detection, is one of the most common quality control types in manufacturing. The task is to identify the presence of an anomaly given an image, e.g., a missing component on an image of a circuit board, for subsequent manual inspection. While industrial anomaly detection has seen a surge in recent years, most anomaly detection methods still utilize knowledge only from normal samples, failing to leverage the information from the frequently available anomalous samples. Additionally, they heavily rely on very general feature extractors pre-trained on common image classification datasets. In this paper, we address these shortcomings and propose the new anomaly detection system AnomalousPatchCore~(APC) based on a feature extractor fine-tuned with normal and anomalous in-domain samples and a subsequent memory bank for identifying unusual features. To fine-tune the feature extractor in APC, we propose three auxiliary tasks that address the different aspects of anomaly detection~(classification vs. localization) and mitigate the effect of the imbalance between normal and anomalous samples. Our extensive evaluation on the MVTec dataset shows that APC outperforms state-of-the-art systems in detecting anomalies, which is especially important in industrial anomaly detection given the subsequent manual inspection. In detailed ablation studies, we further investigate the properties of our APC.


\end{abstract}
\section{Introduction}
\begin{figure*}[ht!]
    \centering
    \begin{subfigure}{0.24\textwidth}
        \includegraphics[width=\textwidth]{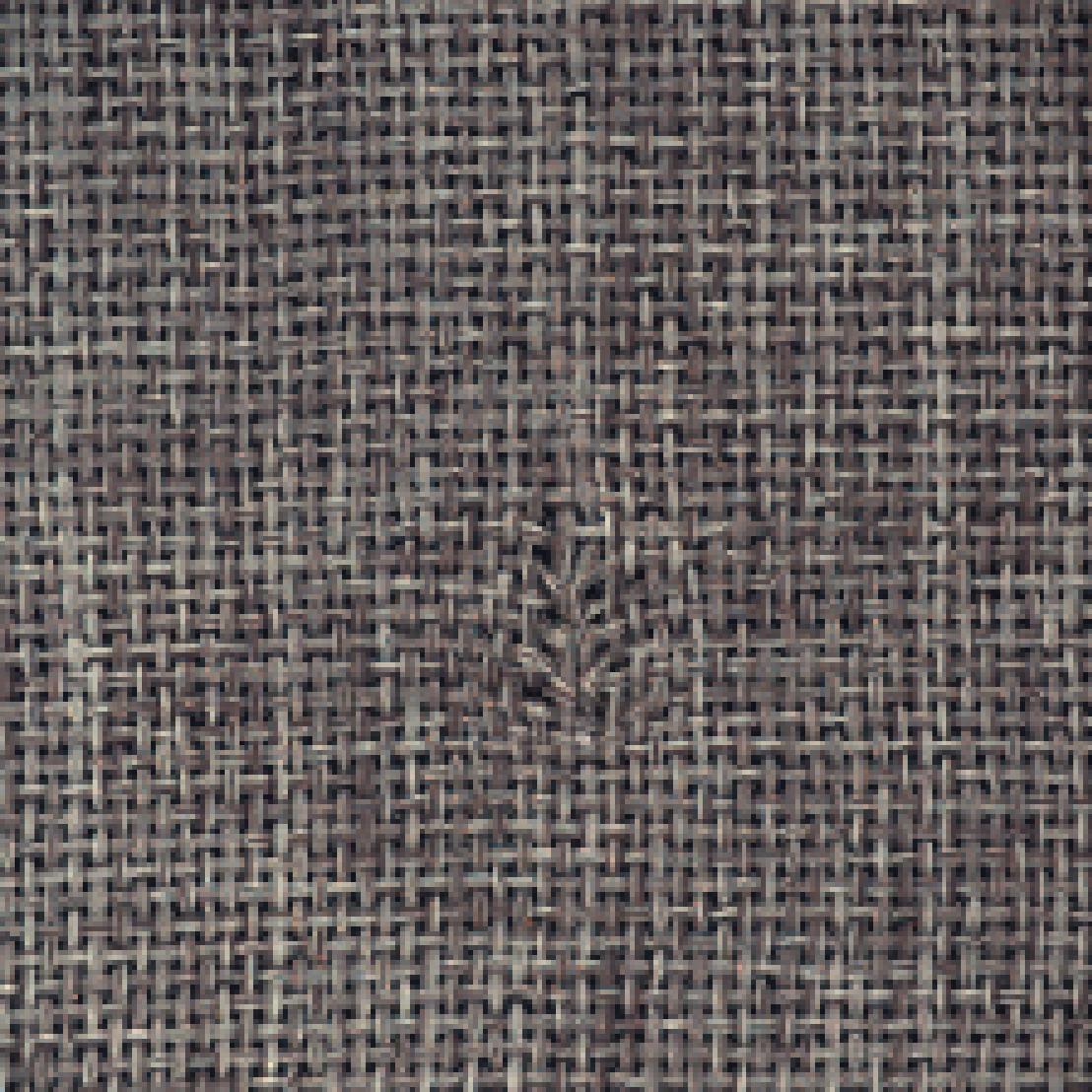}
        \caption{Input}
    \end{subfigure}\hfill
    \begin{subfigure}{0.24\textwidth}
        \includegraphics[width=\textwidth]{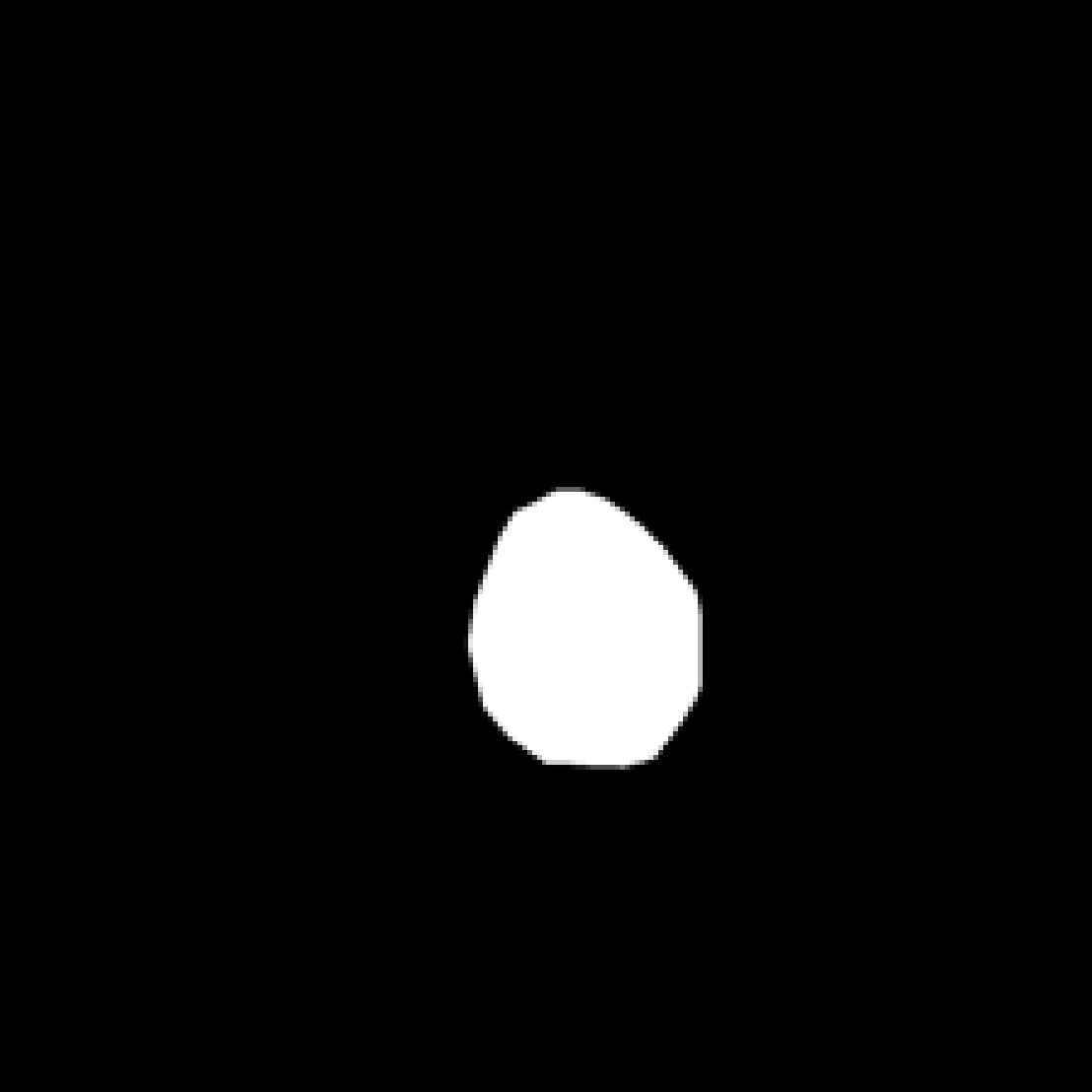}
        \caption{GT segm.}
    \end{subfigure}\hfill
    \begin{subfigure}{0.24\textwidth}
        \includegraphics[width=\textwidth]{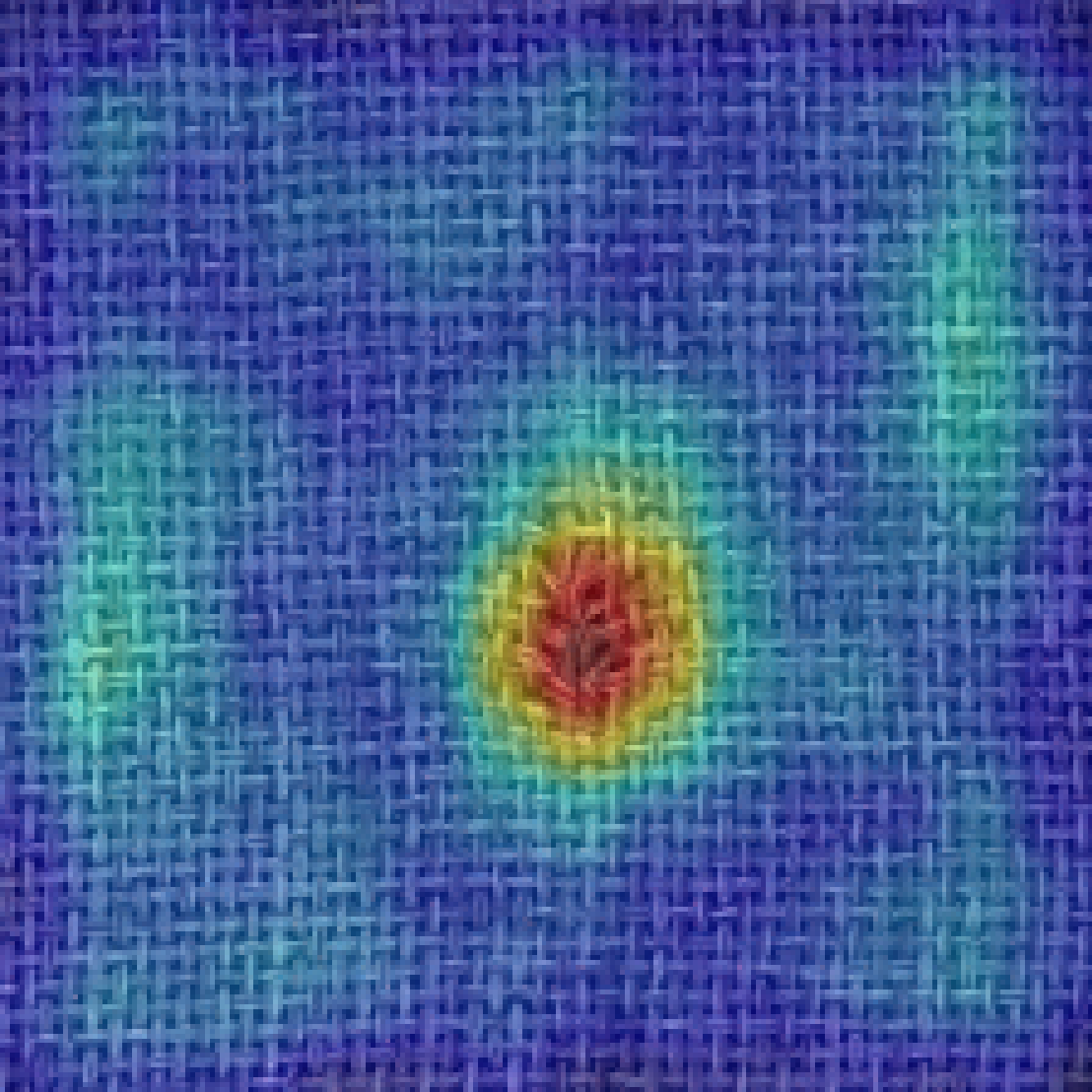}
        \caption{PatchCore~\cite{roth2022patchcore}}
    \end{subfigure}\hfill
    \begin{subfigure}{0.24\textwidth}
        \includegraphics[width=\textwidth]{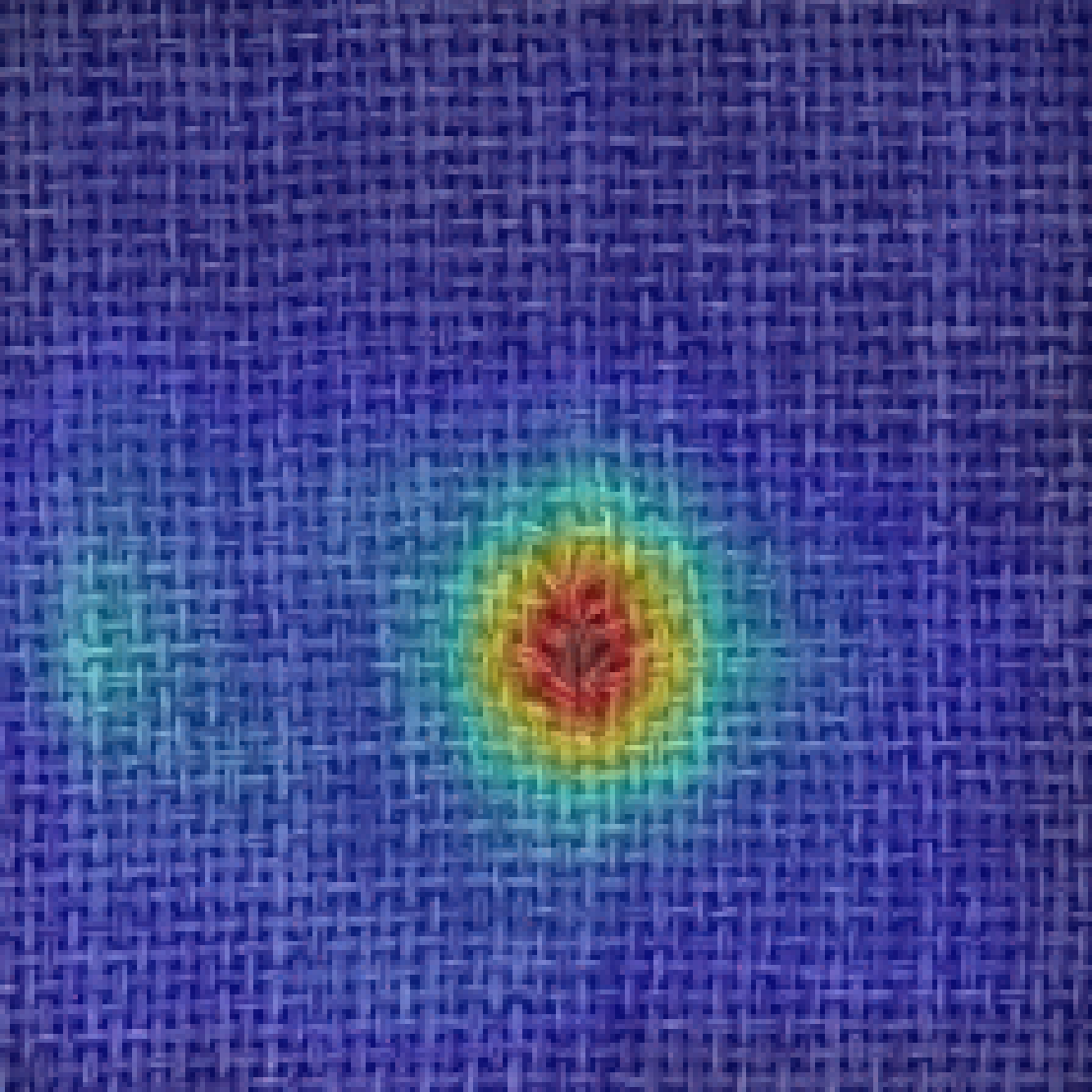}
        \caption{APC (ours)}
    \end{subfigure}
    \caption{Result of our proposed anomaly detection system APC and PatchCore~\cite{roth2022patchcore} on the MVTec dataset. Localization results are overlayed with the original images. Red indicates likely anomalous areas, while blue indicates likely normal areas.}
    \label{fig:mvtec_example}
\end{figure*}

Anomaly detection (AD) is a common task in computer vision. Visual anomalies can manifest as unusual features, unexpected patterns, or distortions within an image or a sequence of images. For example, an anomaly could be an image from a different distribution than normal samples. Given the high relevance of visual AD in manufacturing, a whole sub-field called industrial anomaly detection has emerged~\cite{ind_ad_survey}. Due to the controlled nature of the manufacturing process, what separates it from other application domains is a more constrained problem formulation. For instance, in Fig.\ref{fig:mvtec_example} the background is usually uniform, without distractors and lightning artifacts.

In the present literature, most work in the field of industrial anomaly detection is set up in a classical one-class classification (OOC) setting~\cite{ind_ad_survey}, where the model is trained to detect anomalies using only samples from the one class, in this case, normal samples. This motivation stems from the infrequency of anomalous samples. Since the range of possible anomalies is huge, it is infeasible to get a dense sample from the distribution of all possible anomalies with few anomalous samples. Thus, supervised learning is inapplicable. 

However, obtaining a few anomalous samples is possible in most use cases~\cite{liu2019margin}. For example, quality control systems allow retrieving false-negative detections discovered after a manual inspection. This gap is currently being addressed, with an increasing amount of work operating in semi-supervised OOC settings~\cite{one_class_class_survey}, meaning the utilization of anomalous samples during training, e.g.,~\cite{yao2023clflow,zhang2023prototypical,Li2023EfficientAD,Wei2024FewshotOA,Wang2024WeaklySA}. Most of these methods have complicated pipelines and heavily rely on synthetically generated anomalies~\cite{zhang2023prototypical,yao2023clflow} for training. 

In this paper, we also investigate using anomalous samples for industrial anomaly detection. Specifically, we propose a new, simple anomaly detection system that utilizes both normal and anomalous samples for training in a supervised fashion. Our novel system, AnomalousPatchCore~(APC), is based on PatchCore~\cite{roth2022patchcore}, however, the idea applies to any approach based on feature embeddings. APC enhances the pre-trained feature extractor in PatchCore, pre-trained on ImageNet, by incorporating information from anomalous samples. Specifically, APC leverages the information from these anomalous samples by fine-tuning the feature extractor to generate features that better fit the anomaly detection task compared to ImageNet-based features. To achieve this, we introduce three new auxiliary training tasks to the  feature extractor's architecture, while keeping the later parts of PatchCore. Our auxiliary training tasks cover classification, segmentation, and reconstruction to leverage anomalous samples effectively without additional annotations. Our extensive evaluation on the MVTec dataset highlights the superior detection performance of APC compared to PatchCore. Moreover, the evaluation also shows the large potential of the extracted features and gives further insights into the properties of APC through four ablations studies.




To sum up, the major contributions of this paper are the following: 
\begin{itemize}
    \item We propose APC, a new and simple anomaly detection method that effectively leverages knowledge from a few anomalous samples by improving the feature extraction.
    \item To leverage the knowledge of anomalous samples, we design three auxiliary tasks and prove their effectiveness through ablation studies.
    \item Our comprehensive evaluation shows the benefit of using APC with anomalous samples for detecting anomalies compared to state-of-the-art systems trained on normal samples only.
\end{itemize}




\section{Related work}
Anomaly detection methods can be divided into unsupervised, if only normal samples are available, and supervised, if the training set contains normal and anomalous samples~\cite{ind_ad_survey}.  

\subsubsection{Unsupervised Anomaly Detection}
Unsupervised methods are divided into feature-embedding, reconstruction-based, and gradient-based methods. Feature-embedding methods  first extract features employing a pre-trained model and, during inference, compare the extracted features of a sample with stored normal ones to identify unusual features, i.e., anomalies. A main difference between feature-embedding systems is how extracted features are handled. While many systems utilize a memory bank of features~(samples)~\cite{cohen2021subimage,liu2023simplenet,roth2022patchcore} and determine the closest feature during inference, others fit distributions on the extracted features and perform a statistical test during inference to determine whether a feature comes from the distribution of normal features~\cite{Defard2020PaDiMAP,sohn2021learning,Yang2022MemSegAS}. Finally, \cite{Tien_2023_CVPR,Deng2022AnomalyDV} use knowledge distillation and compare features extracted by a teacher and student to perform anomaly detection. 

Reconstruction-based methods for anomaly detection aim to reconstruct the input. If the reconstruction is inaccurate, a sample is regarded anomalous. \cite{Xia2022GANbasedAD}~use a generative adversarial network for the reconstruction with a generator that learns to produce synthetic normal samples, while the discriminator distinguishes between real and generated samples. In contrast, \cite{baur2019deep,Gudovskiy2021CFLOWADRU,yao2023clflow} utilize variational autoencoders or normalizing flows to map each sample to a latent space and reconstruct it with a decoder. During inference, the models accurately reconstruct normal samples, while anomalous samples produce larger errors. Gradient-based methods exploit the assumption that anomalous samples often exhibit higher variations in feature gradients compared to normal ones. Analyzing the magnitude or pattern of the gradients makes it possible to identify anomalous instances~\cite{lee2022gradientbased,kwon2020backpropagated}. 

\subsubsection{Supervised Anomaly Detection}
Supervised methods leverage anomalous samples to perform anomaly detection, yet, few methods exist in this category, especially for industrial settings. For example, \cite{Liznerski2020ExplainableDO}~utilize specific loss terms to leverage labeled samples, while \cite{Ding2022CatchingBG} train an encoder to produce disentangled representations of anomalies. 
In contrast, \cite{zhang2023prototypical} propose an attention-based U-Net utilizing prototypes, i.e., representations of average normal samples created by clustering. To counter the class imbalance between normal and anomalous samples, \cite{zhang2023prototypical} utilize a sophisticated pipeline for generating anomalous samples.

We generally follow the feature-embedding paradigm of the unsupervised methods and especially the idea of~\cite{roth2022patchcore}. However, different from these methods, we fine-tune the feature extractor utilizing normal and anomalous samples based on three auxiliary tasks to improve the discrimination between these samples during inference. Hence, we bridge the gap between the unsupervised feature-embedding paradigm and supervised methods, without relying on sophisticated data generation.

\section{Baseline}
\label{sec:patchcore}

\begin{figure}
    \centering
        \includegraphics[width=0.85\linewidth]{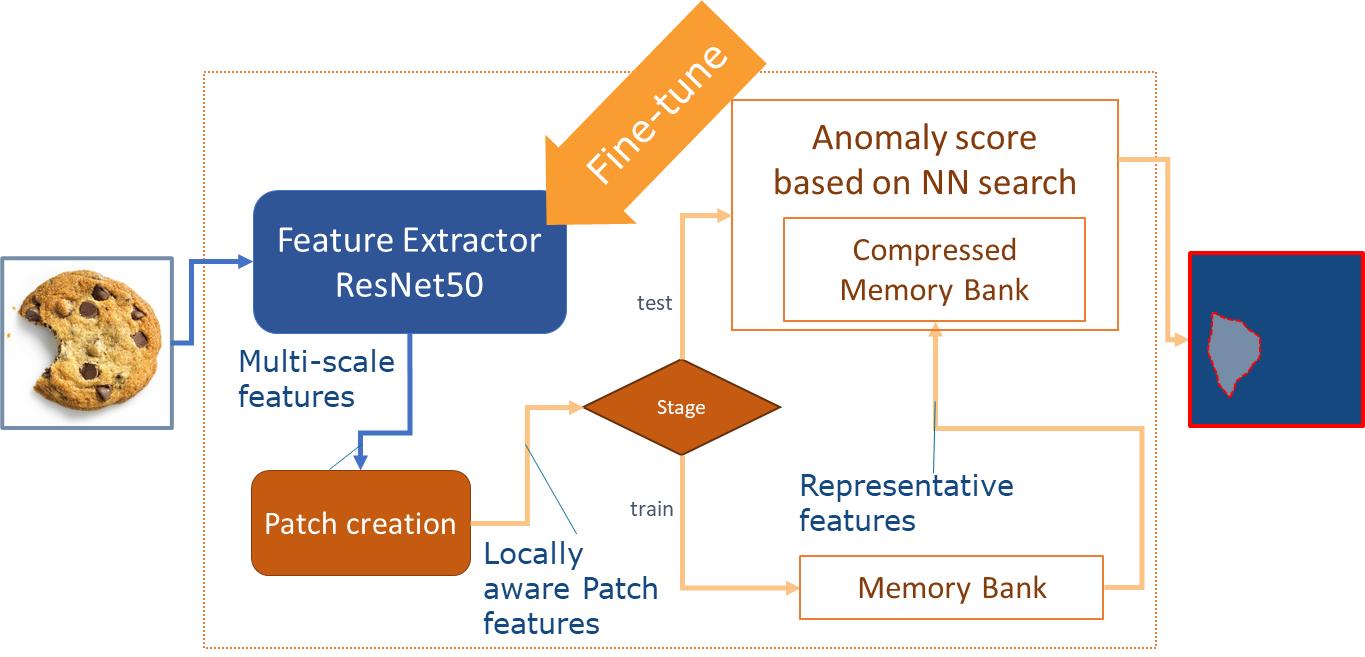}
    \caption{Abstract system figure of baseline PatchCore with three major components: the feature extraction~(top left), the patch creation~(bottom left), and the memory bank~(right side). In our proposed system APC~(see \cref{sec:method}), we improve the feature extractor marked by the arrow, while keeping the rest of PatchCore.}
    \label{fig:patchcore}
\end{figure}

In this section, PatchCore~\cite{roth2022patchcore} shown in Fig.~\ref{fig:patchcore} is introduced, since our novel anomaly detection system APC, described in \cref{sec:method}, is based on this model. Note that APC is also easily applicable to other feature-embedding approaches. As with other feature-embedding methods, PatchCore comprises three key stages.  

The first stage is the feature extraction~(top left in \cref{fig:patchcore}) from normal samples of the training set. It is done using a feature extractor, e.g., a ResNet-50 pre-trained on a general classification dataset like ImageNet. Note that in PatchCore, this feature extractor is not fine-tuned.
As the feature extractor performs forward propagation, each layer of the feature extractor produces representations of the image with increasingly richer semantic information. In PatchCore, only the representations of two intermediate layers are used~(for ResNet-50 the output of stage 3 and stage 4). This follows the idea that the best features for anomaly detection are the ones that are not too specific to the dataset on which the feature extractor was trained, and not too low-level. Subsequently, the features extracted from the two ResNet-50 layers are up-scaled to the resolution of the larger one and concatenated, to combine features over multiple semantic levels.

Achieving some degree of translation invariance, PatchCore extracts patches from the combined feature map, which is the second stage of PatchCore~(bottom left in \cref{fig:patchcore}). To create patches, the combined feature map is processed by aggregating the features over the neighborhood of a pre-defined size by average pooling. 
The resulting locally aware patch features are stored in a memory bank of only normal samples, which is the third stage in PatchCore~(bottom right in \cref{fig:patchcore}). Finally, to speed up the inference, representative samples in a predefined ratio are selected following the core set procedure.


During inference~(top right in \cref{fig:patchcore}), features are extracted from a test sample using the same feature extractor as in training, and, subsequently,  test patches are created. 
Assigning an anomaly score is done by measuring the distance of each test patch to the closest patch from the memory bank created during training. 
To obtain the anomaly map, the score of each patch is distributed over the neighborhood used during feature aggregation. 
The overall image score is the maximum anomaly score among the patches.

\section{Method}
\label{sec:method}

\begin{figure}
    \centering
        \includegraphics[width=0.66\linewidth]{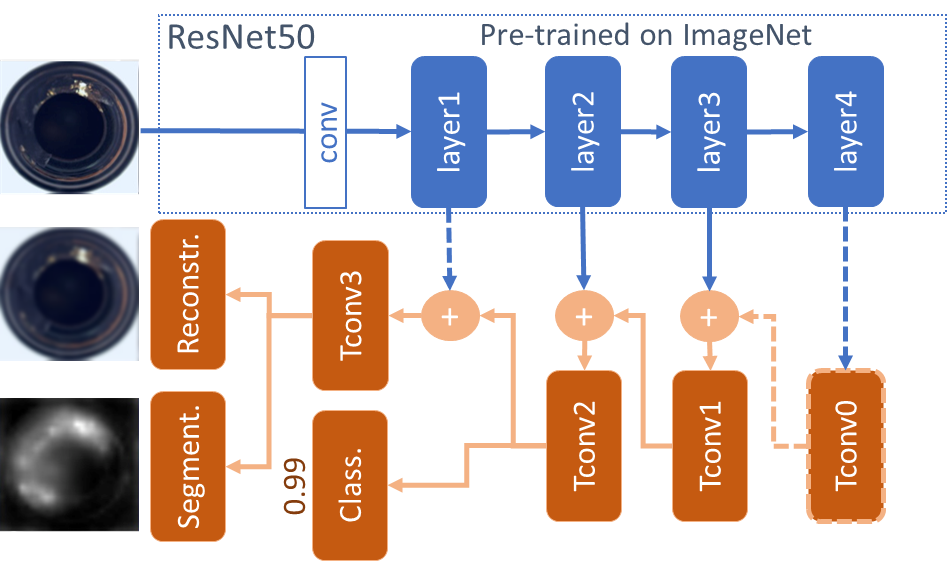}
    \caption{Detailed system figure of the feature extractor in our APC during training. Based on a ResNet-50~(top), we construct a U-Net decoder~(bottom right) with two skip-connections~(optionally four, dashed lines) and three task-specific heads on top~(bottom left). Here, \textit{Tconv} denotes transposed convolution. Each head corresponds to one of the auxiliary tasks: classification, segmentation, and reconstruction. Note that the remainder of the anomaly detection system is similar to PatchCore~(see \cref{fig:patchcore}). }
    \label{fig:sysFig}
\end{figure}

This section introduces our new and simple anomaly detection system AnomalousPatchCore~(APC), which leverages the knowledge of anomalous samples in a framework similar to PatchCore~(see \cref{sec:patchcore}). To leverage the knowledge of anomalous samples, we design and train a new feature extractor for anomaly detection based on a ResNet-50, visualized in \cref{fig:sysFig}. For training the feature extractor on both normal and anomalous samples, we extend the ResNet-50 architecture into a U-Net architecture~(bottom right in \cref{fig:sysFig}) and add three auxiliary tasks based on classification, segmentation, and reconstruction~(bottom left in \cref{fig:sysFig}) as described in \cref{sec:method_fe}. Once features are extracted, APC applies a patch-based memory bank-approach similar to PatchCore~(see \cref{sec:meth_anom}). Finally, we provide details of APC's implementation and training in \cref{sec:meth_impl_details}.


\subsection{Feature Extraction}
\label{sec:method_fe}
The core novelty of APC is a new feature extractor that leverages knowledge from anomalous samples by fine-tuning. This way, the memory bank-based anomaly detection operates on features with a larger distance between normal and anomalous regions, improving the difference between normal and anomalous image areas, which boosts anomaly detection. To fine-tune the feature extractor with normal and anomalous samples, we start from the same ResNet-50 feature extractor as PatchCore and add three carefully selected auxiliary tasks, described below, which drive the fine-tuning. Note that we also adapt the initialization of the ResNet-50~(see \ref{sec:meth_impl_details}).

\subsubsection{Classification}

The first auxiliary task for fine-tuning the feature extractor in APC is image-level \textit{classification}. Hence, the feature extractor is trained to generate features for predicting whether a sample is normal or anomalous. This task is naturally derived from the anomaly detection task, which features both, image-level classification and pixel-level localization.  Another advantage of the classification task is the low data labeling costs, requiring only images and their corresponding labels, which is much less labor-intensive compared to segmenting anomalies in images. However, a disadvantage of this task is that it provides less granular information about the anomalies, giving the model less information to train on, and potentially learning the wrong features that may correlate with the anomalies, e.g., Clever Hans predictors.



\subsubsection{Segmentation}
Similar to classification, \textit{segmentation} is the second auxiliary task in APC that naturally arises from the general setup of anomaly detection, resembling pixel-level localization. The segmentation task aims to replicate the pixel-wise classification into normal and anomalous. Hence, segmentation is a binary segmentation task. Different from classification, the segmentation task forces the network to attend to the relevant parts of the image~(anomalies) and, therefore, produces features that better discriminate between normal and anomalous parts. Additionally, as the model naturally learns to identify anomalies, it's compelled to understand the context and relationships between objects for anomaly detection. This understanding should lead to the production of spatially aware features by the feature extractor, ultimately achieving the desired outcome of fine-tuning. As visible in \cref{fig:sysFig}, the segmentation is implemented by extending the ResNet-50 to a U-Net~\cite{Ronneberger2015UNetCN} with two skip connections as visible in the bottom right of \cref{fig:sysFig}. We use two skip connections here since the subsequent parts of APC, similar to PatchCore, only use the features from these two layers of the ResNet. We validate this choice in \cref{sec:eval_abl}. On top of the U-Net, we add a head to predict the binary segmentation.


\subsubsection{Reconstruction}
The final auxiliary task in APC is a \textit{reconstruction} of the original input image. Given the limited availability of negative samples yielding a substantial class imbalance between negative and positive samples, the reconstruction task can effectively mitigate this imbalance by learning relevant features for anomaly detection from both types of samples. In this task, the model is trained to reproduce the input image from its compressed latent representation, i.e., embeddings. To implement this task, we add another head to the U-Net architecture dedicated to the reconstruction task. 



\subsection{Detecting Anomalies} 
\label{sec:meth_anom}
Besides the novel feature extractor trained with the auxiliary tasks described in \cref{sec:method_fe}, APC is similar to PatchCore. Hence, after training the feature extractor with the new auxiliary tasks, APC extracts local patches based on the concatenated feature maps of the feature extractor. Similar to PatchCore, stage 3 and stage 4 of the feature extractor's ResNet-50 are used. Note that the decoder part of the feature extractor is not used as this stage. Given the locally aggregated patch features, the same type of memory bank is constructed as in PatchCore. However, the patches of anomalous and normal samples are expected to be further apart in feature space due to the improved feature extraction in APC. Inference is analogous to PatchCore, hence, the ResNet-50 part of our new feature extractor is used to generate the feature representation per image.

\subsection{Implementation Details} 
\label{sec:meth_impl_details}



\subsubsection{Architecture and Training}
As briefly described above, we augment the feature extractor in APC, a ResNet-50, with a U-Net-style decoder. The decoder reconstructs embeddings of the first three stages of the ResNet-50 encoder with transposed convolution layers (\textit{Tconv}) for upsampling supported by skip connections between the respective layers of the encoder and the decoder. On top of the decoder, three heads are added for the three auxiliary tasks to train the feature extractor. For classification, we use the features from \textit{Tconv2}~(see \cref{fig:sysFig}), since this allows the classification task to influence all features used in the patch extraction of APC during training. However, further upsampling does not benefit the classification task, unlike the other two tasks. The classification head itself consists of three layers with convolution, activation, and max-pooling, followed by two fully-connected layers. For training, we use binary cross-entropy loss~($L_{cls}$) between the predicted class and the ground truth label. 

For the auxiliary tasks segmentation and reconstruction, we add the transposed convolution layer \textit{Tconv3}~(see \cref{fig:sysFig}) to upsample the spatial resolution of the feature map for these pixel-level tasks. To facilitate the learning of robust features, the segmentation and reconstruction heads share the \textit{Tconv3} layer. The segmentation head receives the output of the \textit{Tconv3} layer and consists of one more transposed convolution, Gaussian smoothing for smoothed segmentation maps, and convolution followed by sigmoid activation to classify each pixel as anomalous or normal. For training, we add pixel-wise binary cross-entropy loss~($L_{segm}$) between the predicted and the ground truth segmentation maps to the model loss. The reconstruction head comprises transposed convolution, followed twice by a combination of Gaussian smoothing, convolution, and LeakyReLU. The optimization objective for the reconstruction task is the mean squared error~($L_{recon}$) between the input image and its reconstruction. 

Overall, the novel feature extractor is fine-tuned with the three losses described above using AdamW~\cite{loshchilovdecoupled} optimizer for 50 epochs. This step is expensive with respect to the training time, with APC taking around 13 min and PatchCore around 1 min to train on the NVIDIA A5000 GPU, but still acceptable in the most use-cases. To balance the losses, we use the following loss weights, which were established empirically, to calculate the overall loss $L$:
\begin{equation}
    L = 10 \cdot L_{cls} + L_{segm} + L_{recon}.
\end{equation}
As initialization, we use the self-supervised learning MoCo v3 pre-training~\cite{mocov3} on ImageNet to further improve the feature representation for anomaly detection.


\subsubsection{Data Imbalance}
To mitigate the issues caused by data imbalance between normal samples and sparse anomalous samples, we employ two techniques: \textit{balanced sampling} and \textit{data augmentation}. In the presence of a strong class imbalance in the training dataset, the model may tend to reach a suboptimal solution, as the contribution of anomalous samples to the loss function is minimal compared to that of normal samples. Consequently, simply classifying all inputs as normal can already result in a low loss for the model. This issue can be addressed by balanced sampling, which involves sampling normal and anomalous samples with equal frequency by oversampling the anomalous samples. This technique, known as random over-sampling, has been shown to work surprisingly well~\cite{Khushi2021ACP}.

CutPaste augmentation is a more advanced and prevalent approach in visual anomaly detection~\cite{yao2023clflow}, which involves the generation of synthetic anomalies. Following \cite{Li2021CutPasteSL}, a simple and effective way to create anomalies is to take the known anomaly from one image and paste it onto the target image. We employ the implementation from~\cite{zhang2023prototypical}, which combines both basic augmentations with synthetic anomaly generation by applying basic techniques like varying the saturation of the pasted anomalies. These methods enhance the model's ability to learn robust features and improve its performance in detecting anomalies.

\section{Evaluation}

To assess the quality of APC, we evaluate our method on the  MVTec AD\cite{MvTec_ad}, which is a publicly available dataset widely used for evaluating anomaly detection  models~\cite{roth2022patchcore,zhang2023prototypical,liu2023simplenet}. It focuses on the industrial setting and contains 4,096 normal and 1,258 anomalous images close to the ones found in quality control on production. The images are distributed in 15 object/texture categories, e.g., bottles, screws, and leather. Each object/texture category is further divided by the anomaly type present on the image. This dataset is an ideal case for industrial anomaly detection since it is almost noise-free. Each image contains only one object of interest that is precisely positioned in the center of the image on a uniform background without any distractors. 
Following \cite{yao2023clflow},  during fine-tuning, we utilize 10 anomalies in total, sampled uniformly per anomaly type, which can be used to generate synthetic anomalies with CutPaste~\cite{Li2021CutPasteSL}.

To evaluate the performance of our anomaly detection system, we employ four commonly used metrics as highlighted in the survey by \cite{ind_ad_survey}. These metrics provide a comprehensive assessment of the system's effectiveness in both detecting the presence of an anomaly and localizing it within an image. For detecting anomalies at the image level, we use the Area Under the Receiver Operating Characteristic Curve (AUROC$_{im}$) and the F1$_{im}$ score. The AUROC$_{im}$ measures the ability of a model to distinguish between normal and anomalous images by calculating the area under the precision-recall curve for various thresholds. The thresholds here are used on the outcome of an anomaly detection system that is between 0 and 1. For AUROC$_{im}$, higher values indicate better anomaly detection. The F1$_{im}$ score, on the other hand, is the harmonic mean of precision and recall, offering a balance between these two important aspects of performance. The threshold is chosen to maximize the F1$_{im}$. Hence, similar to AUROC$_{im}$, higher values indicate better anomaly detection.  For the localization of anomalies at the pixel level, we use the metrics: AUROC$_{px}$ and the F1$_{px}$ score. Both are defined similarly to the image-level counterparts described above. However, both measure precision and recall based on the level of pixels instead of the image level. Therefore, these measures also assess how well anomalies are localized.


\subsection{Comparison to State-of-the-art on MVTec}
\label{sec:eval_results}
We compare our proposed APC system for anomaly detection with original PatchCore~\cite{roth2022patchcore} using the standard ImageNet pre-trained ResNet-50 feature extractor not utilizing anomalous samples, and PRNet\footnote{We use the \href{ https://github.com/xcyao00/PRNet}{open-source implementation of PRNet}, since no official training code is available. Different from the original paper, this implementation uses a simplified synthetic data generation routine.} \cite{zhang2023prototypical} that utilizes a U-Net architecture similar to APC and strongly relies on synthetically generated anomalous samples. 
Table~\ref{tab:eval_soa} shows the results on the MVTec dataset averaged across the 15 object/texture categories. On the image-level detection task, our APC outperforms both standard PatchCore and PRNet in terms of both AUROC$_{im}$ and F1$_{im}$. While PatchCore is outperformed by 1.62 and 1.51 in AUROC$_{im}$ and F1$_{im}$, respectively, the difference between APC and PRNet is much larger (8.44 and 6.40). The performance gain over PatchCore clearly shows that the proposed refined feature extraction in APC utilizing 10 original anomalous samples per category improves the anomaly detection performance. Hence, even the limited amount of anomalous samples that does not cover the entire range of anomalies, holds valuable information for anomaly detection. Moreover, the results of APC compared to PRNet show that leveraging this information from anomalous samples is possible without substantial synthetic data generation and with only a limited amount of original samples~(here: 10).

\begin{table}[]
\centering 
\caption{Average results across the 15 object/texture categories on the MVTec dataset for our proposed APC, standard PatchCore, and PRNet with 10 original anomalous samples per category.}
\label{tab:eval_soa}
\begin{tabular}{@{}lcccc@{}}
\toprule
System &AUROC$_{im}$ & F1$_{im}$  & AUROC$_{px}$ & F1$_{px}$ \\ \midrule
PatchCore~\cite{roth2022patchcore} &  97.21 & 96.36 & \textbf{97.59} & \textbf{54.70}   \\
PRNet~\cite{zhang2023prototypical} & 90.39 & 91.47 & 93.44 & 54.49  \\
APC (our) & \textbf{98.83} & \textbf{97.87} &  94.45 & 39.04  \\ \bottomrule
\end{tabular}
\end{table}

\begin{figure}
    \centering
        \begin{tabular}{cccc}
             \includegraphics[width=0.24\linewidth]{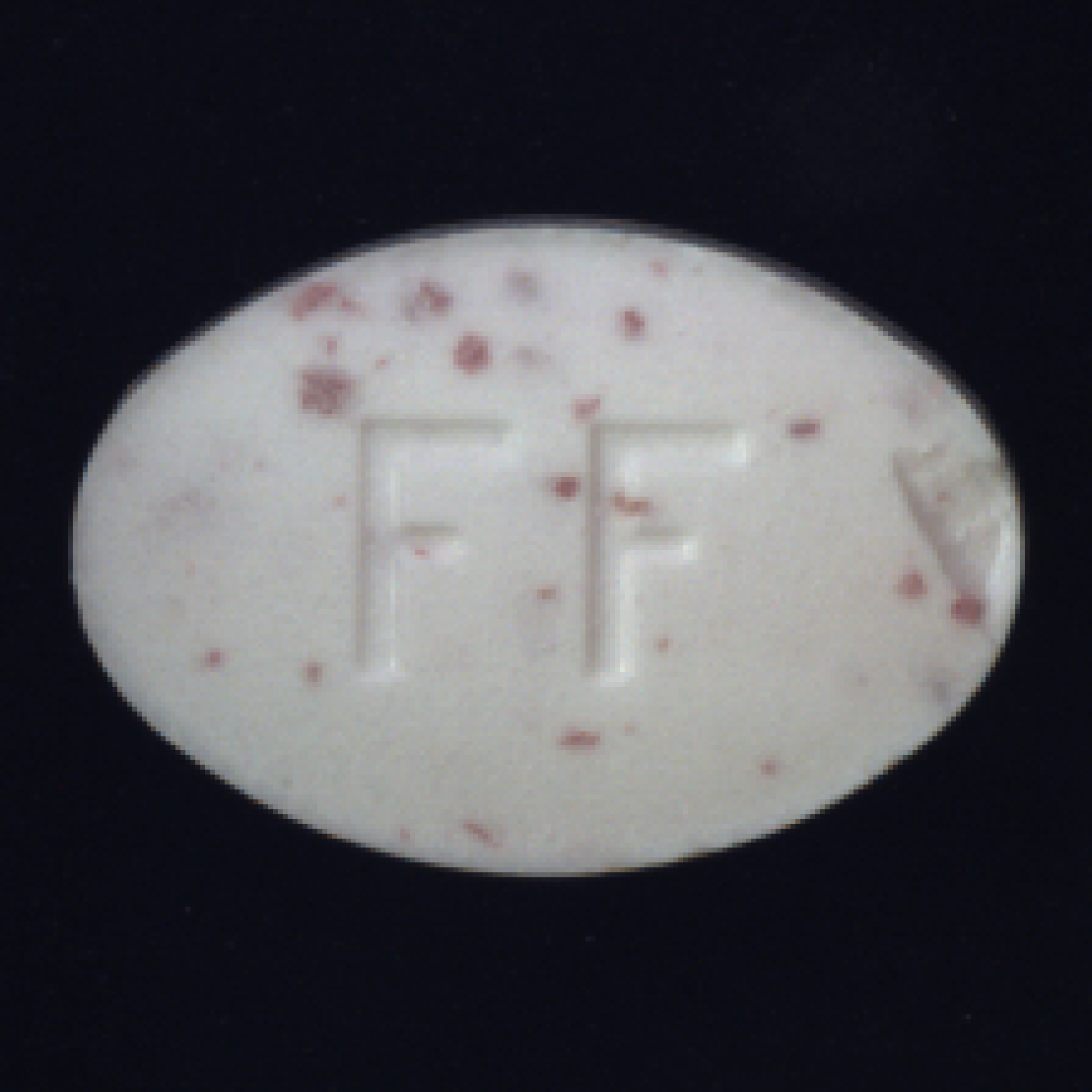}&
             \includegraphics[width=0.24\linewidth]{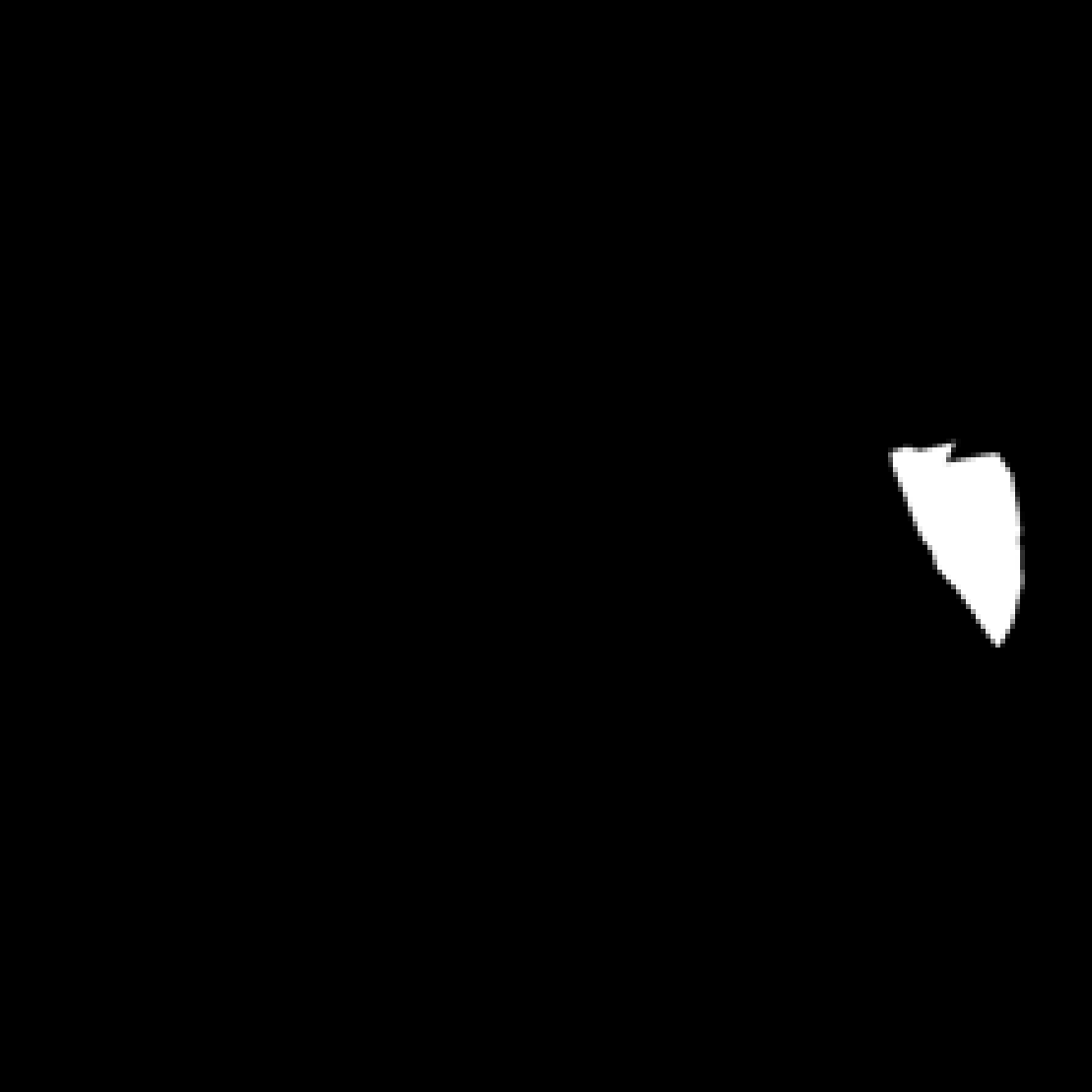}&
             \includegraphics[width=0.24\linewidth]{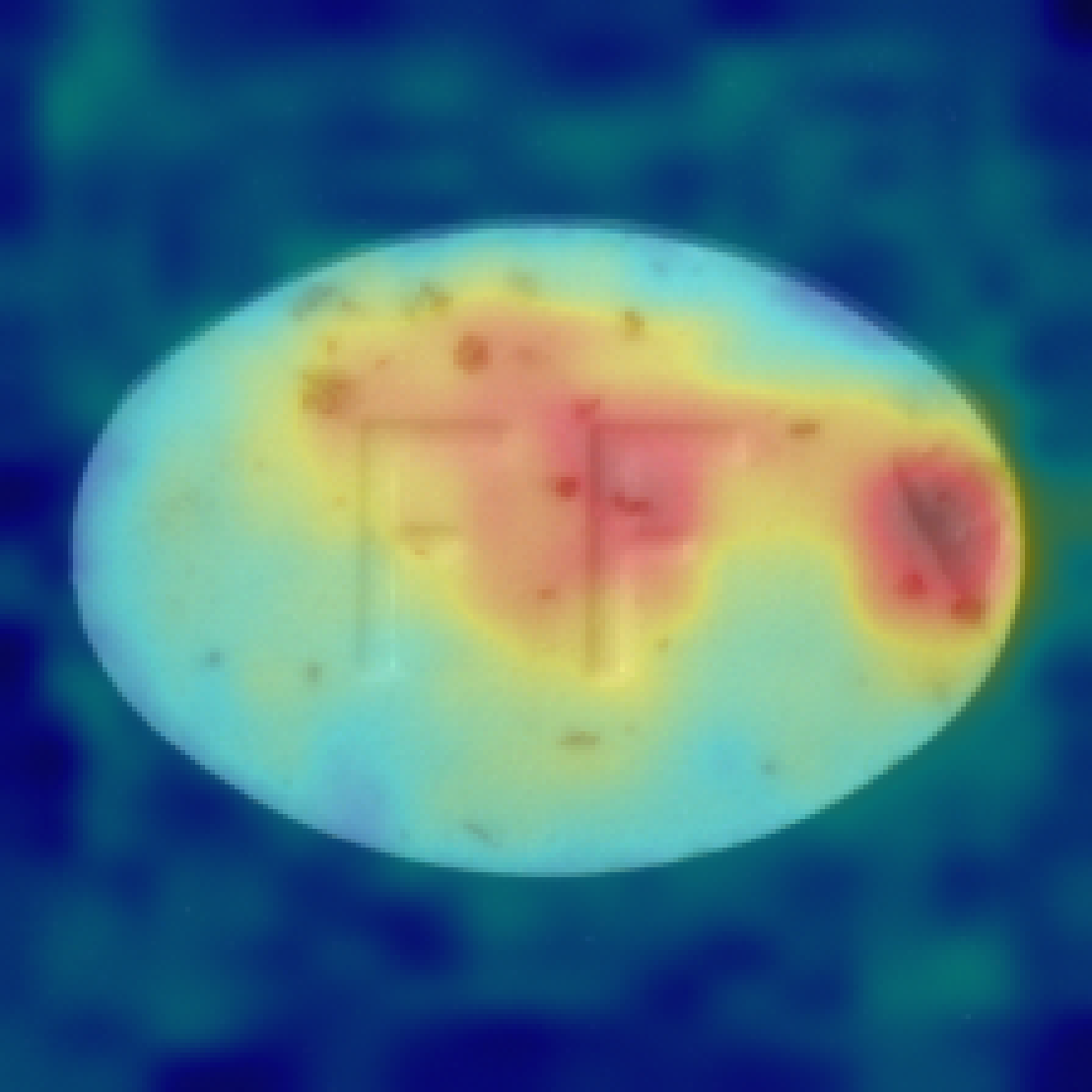}&
             \includegraphics[width=0.24\linewidth]{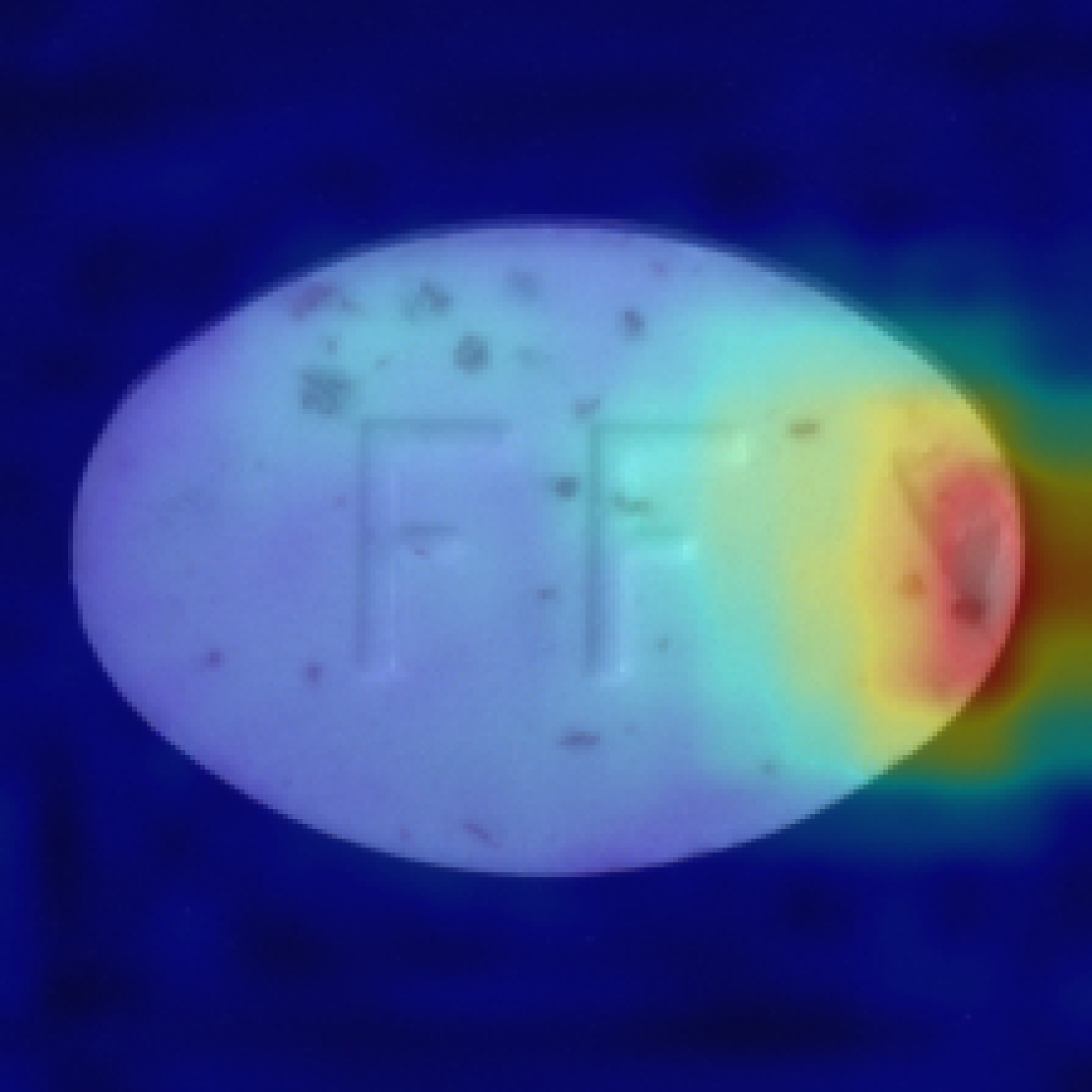}\\ 
             \includegraphics[width=0.24\linewidth]{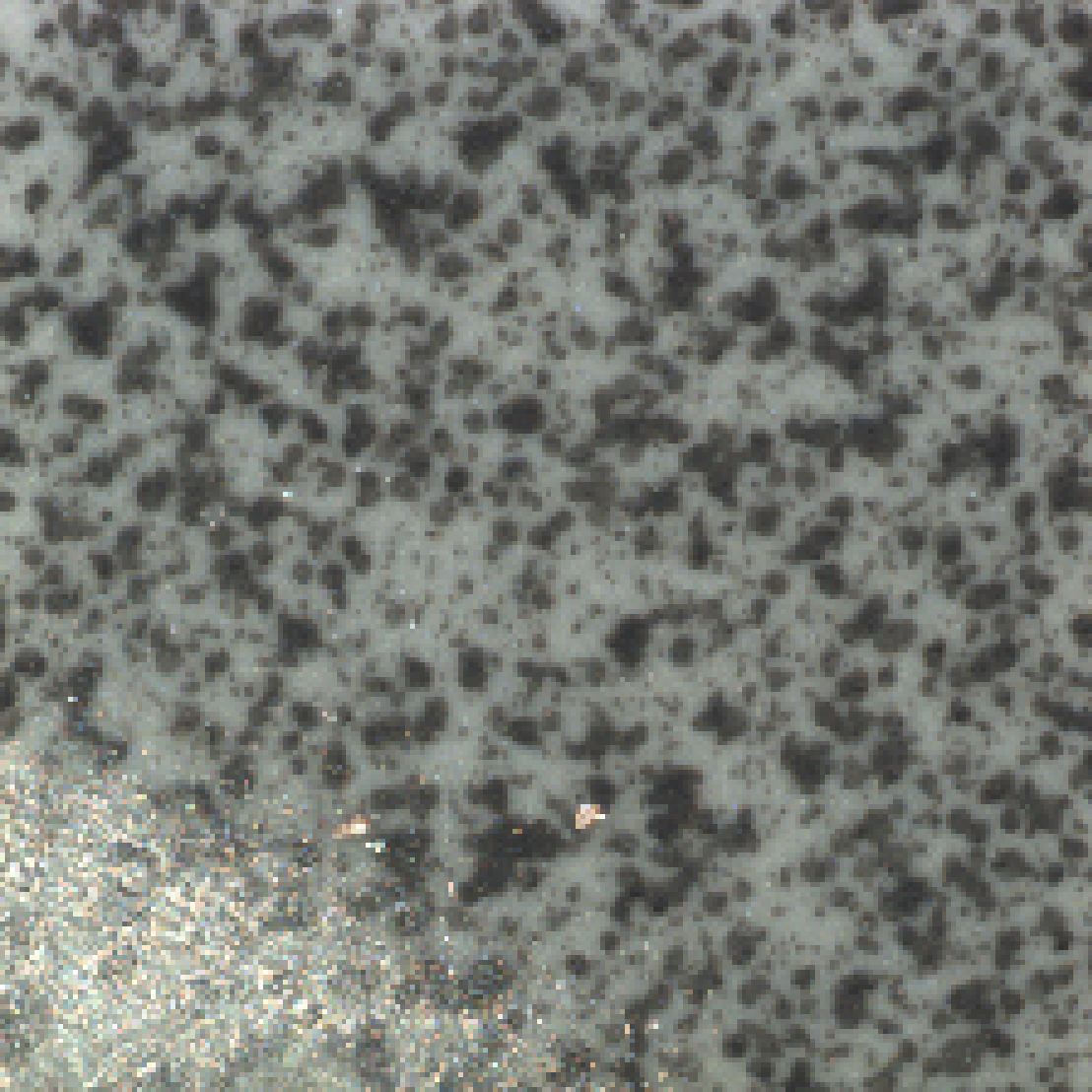}&
             \includegraphics[width=0.24\linewidth]{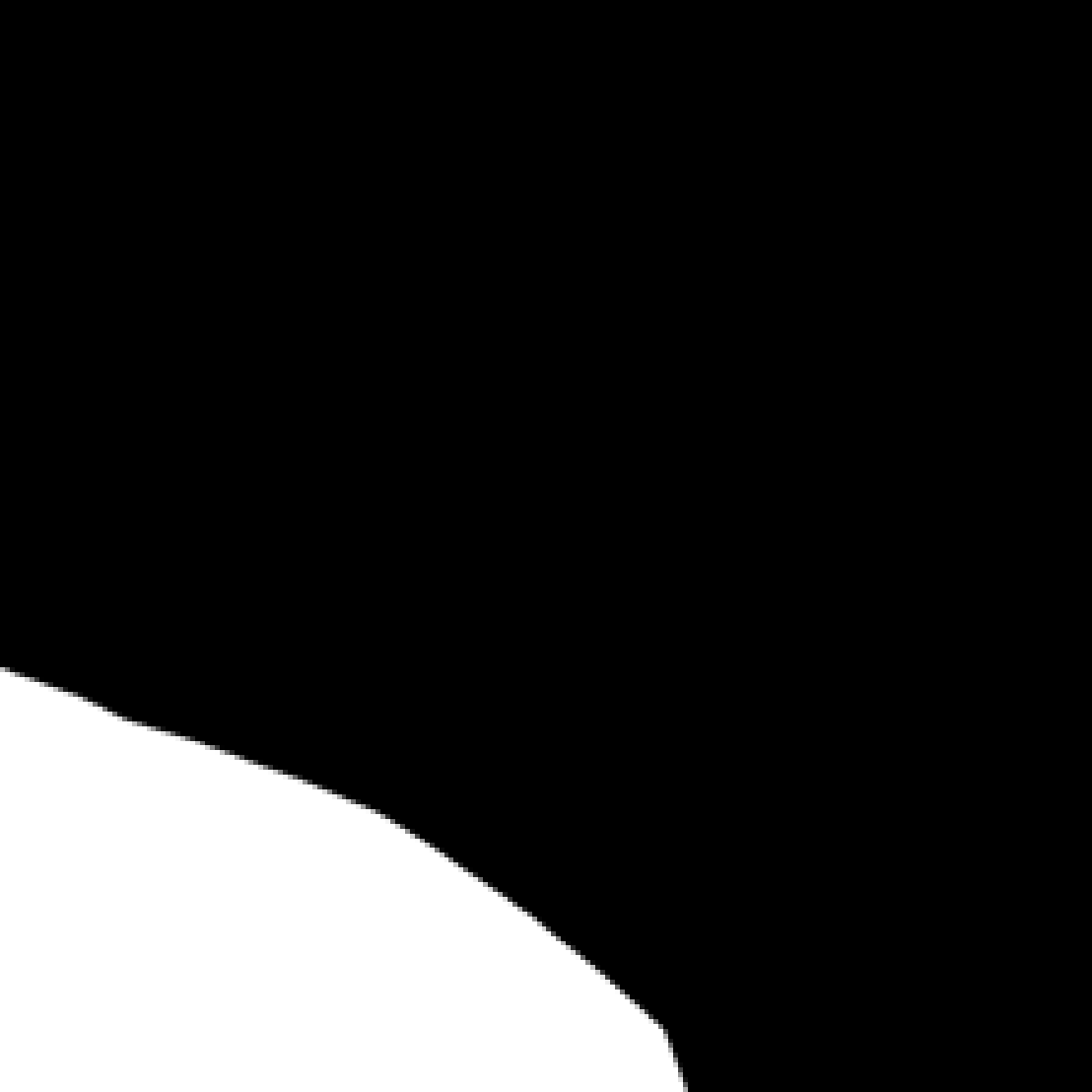}&
             \includegraphics[width=0.24\linewidth]{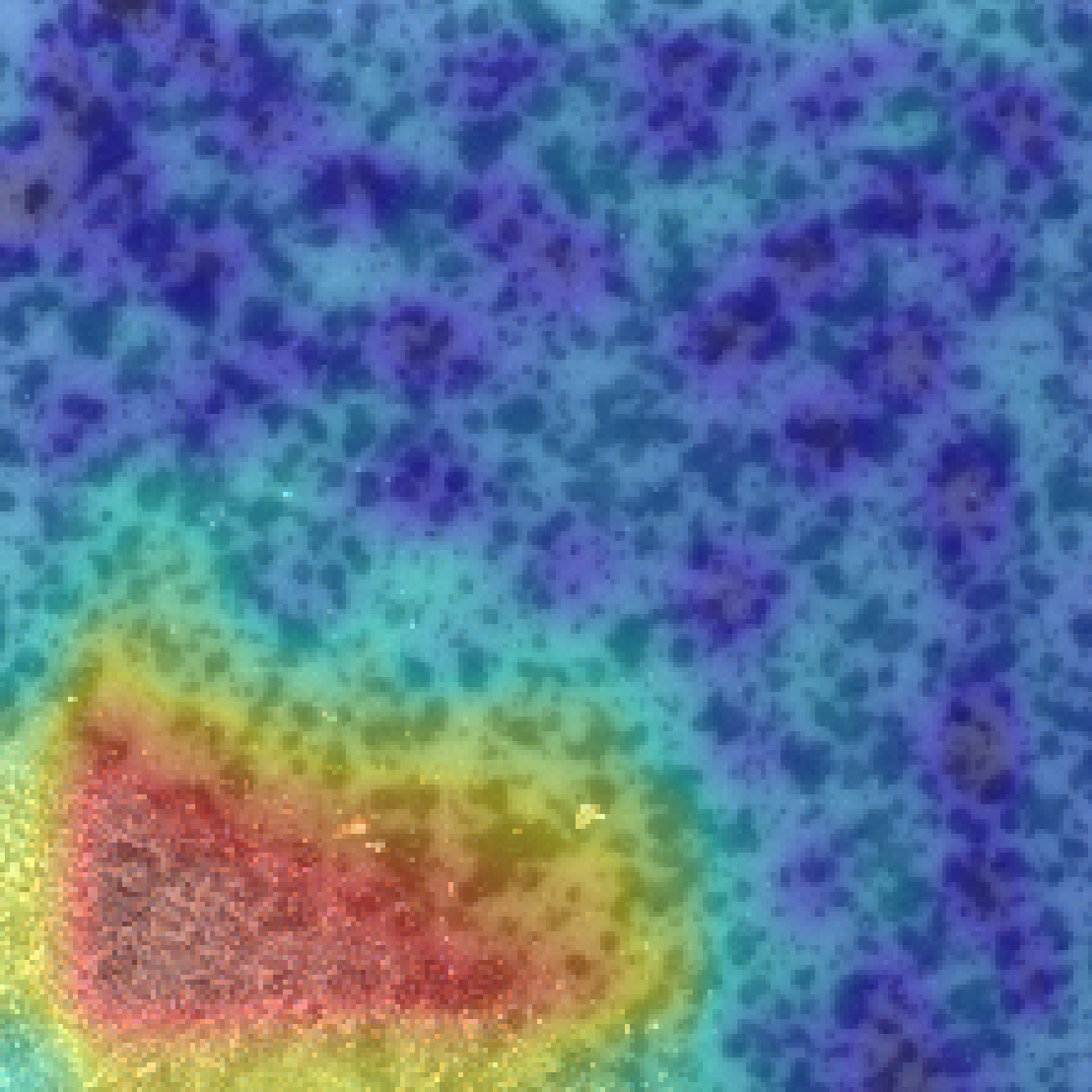}&
             \includegraphics[width=0.24\linewidth]{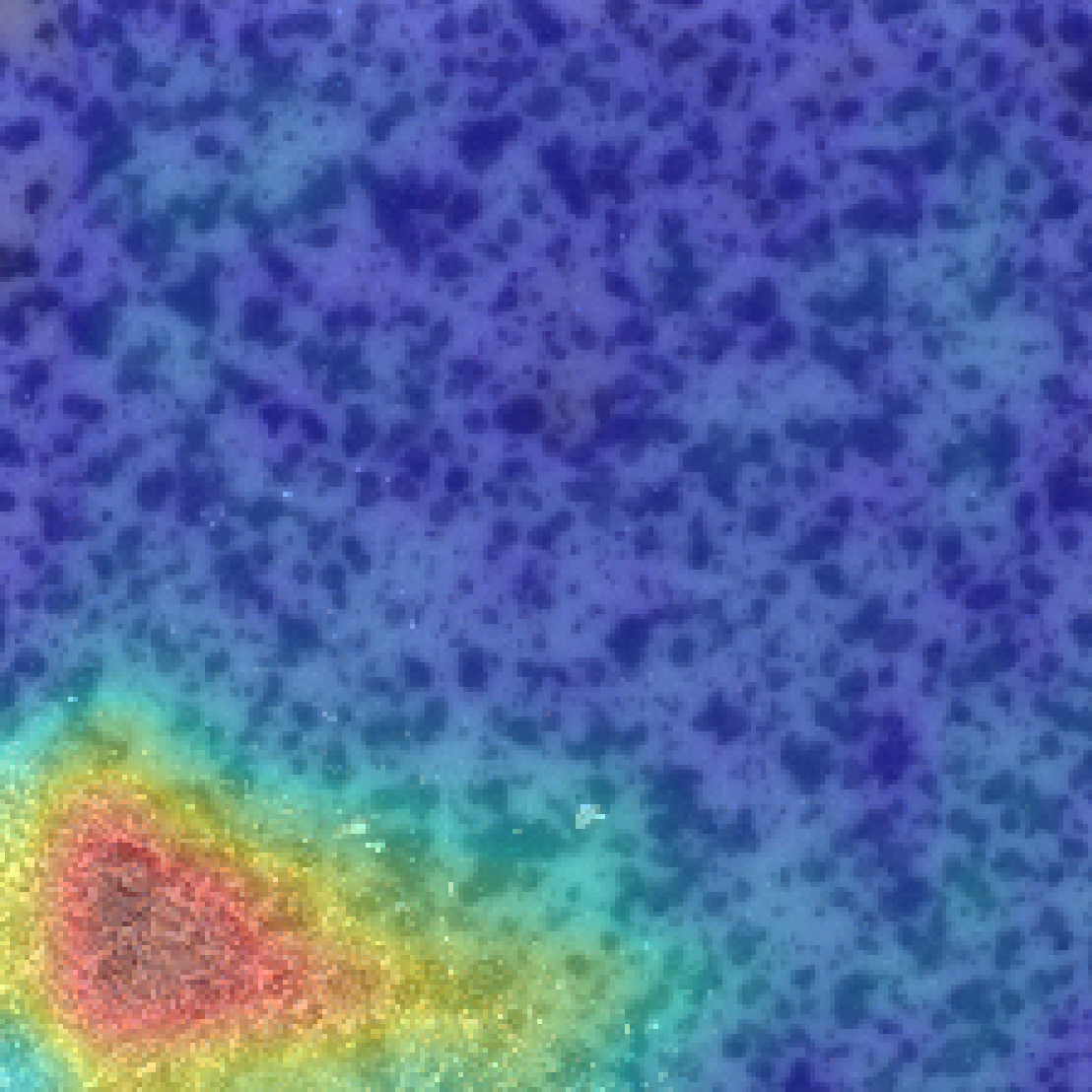}\\ 
             \includegraphics[width=0.24\linewidth]{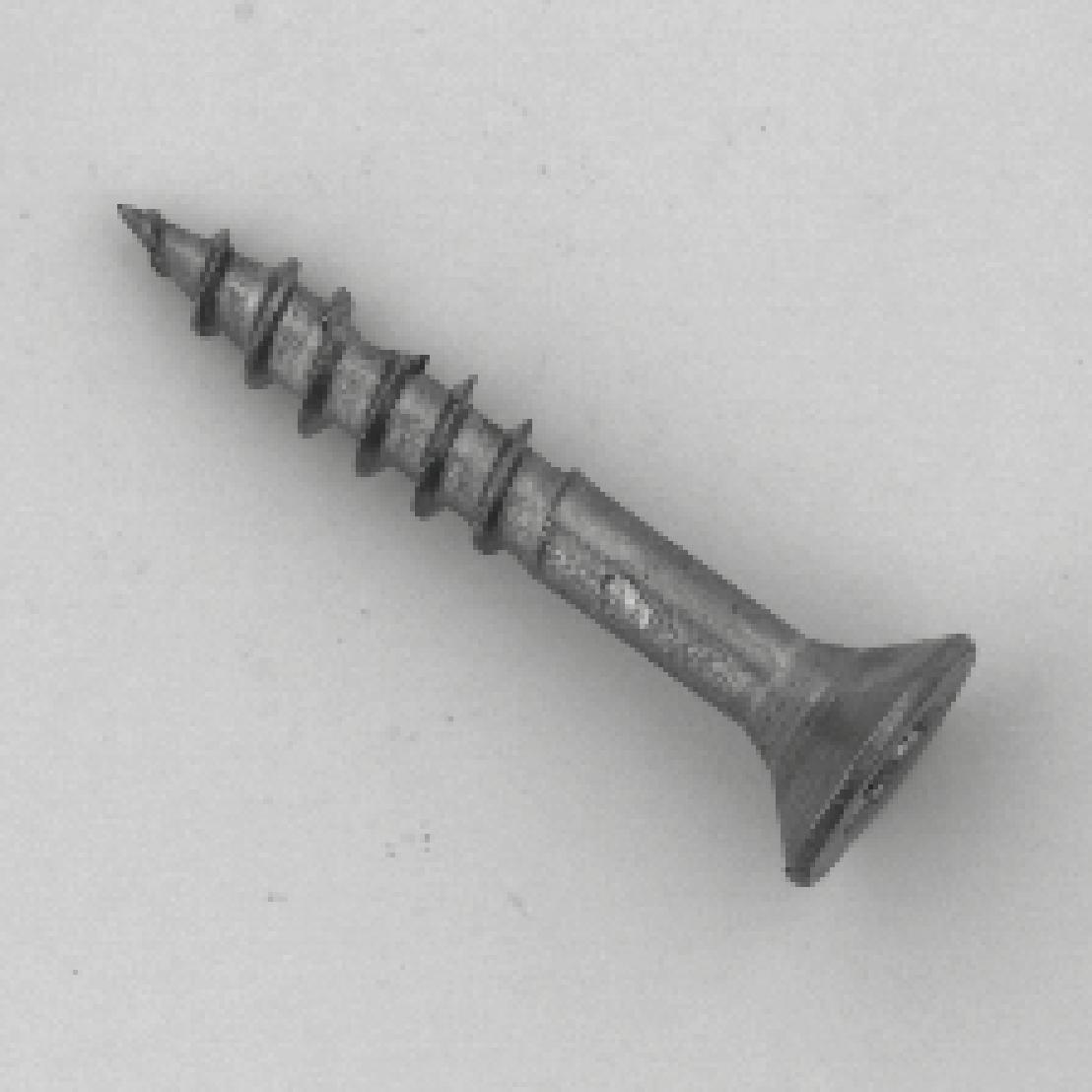}&
             \includegraphics[width=0.24\linewidth]{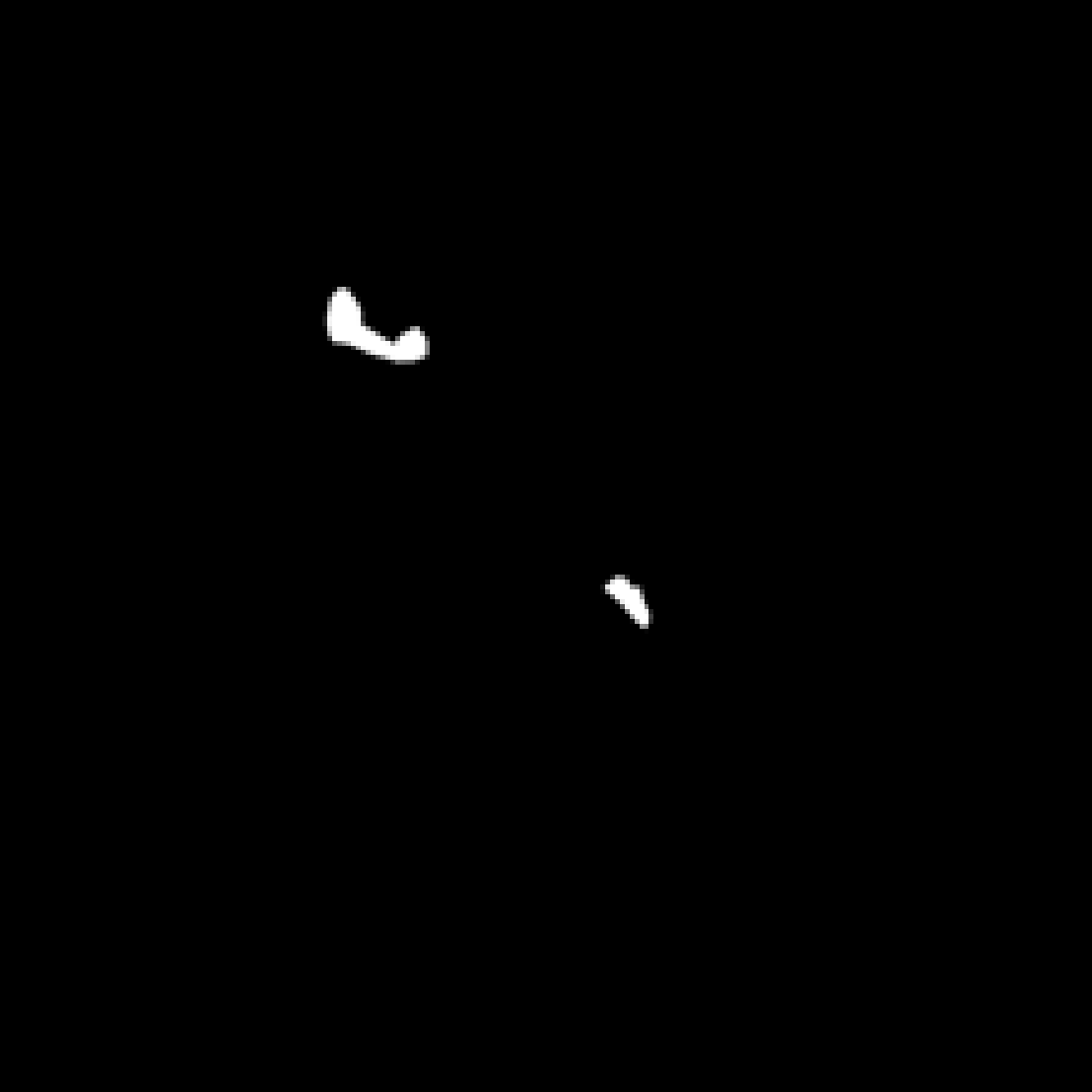}&
             \includegraphics[width=0.24\linewidth]{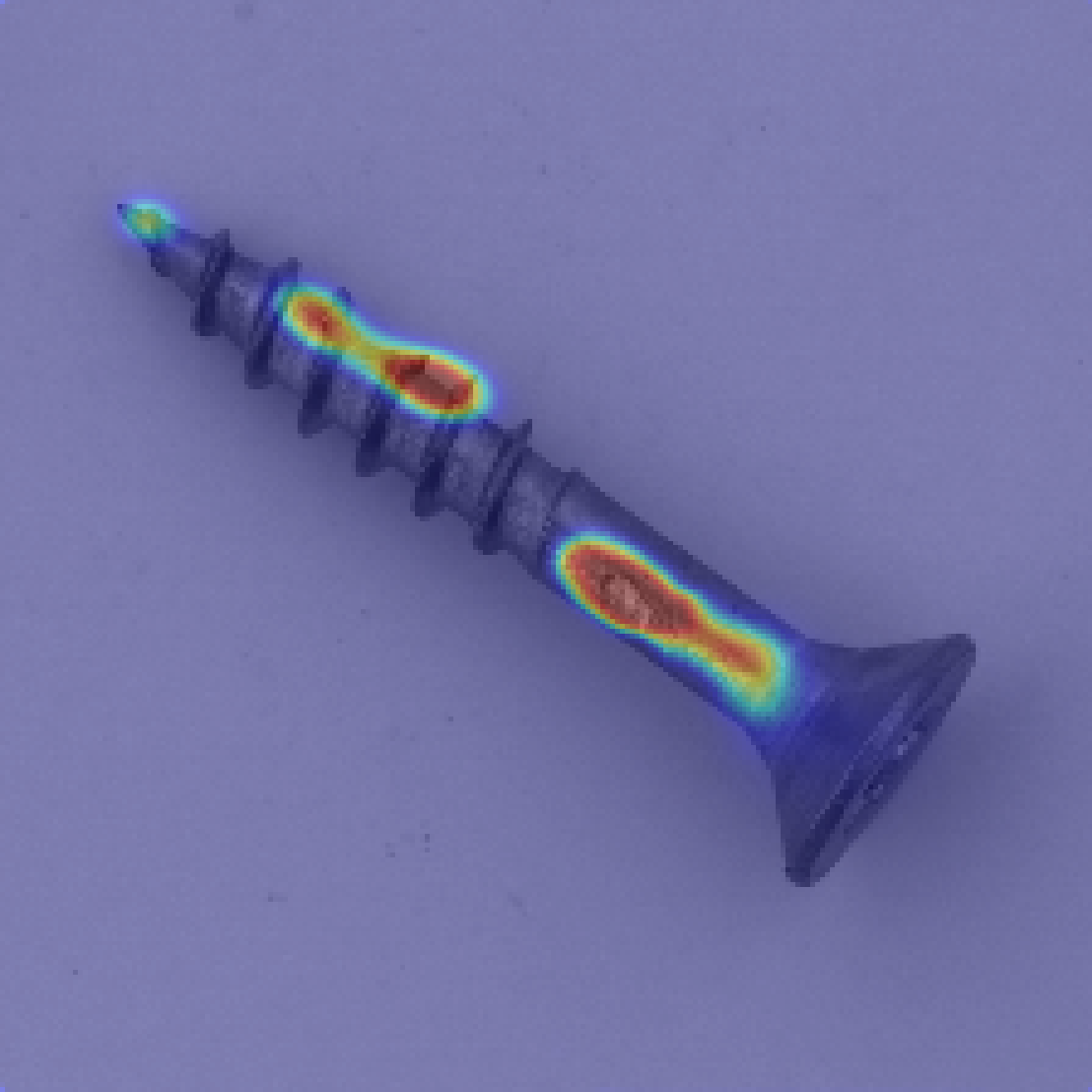}&
             \includegraphics[width=0.24\linewidth]{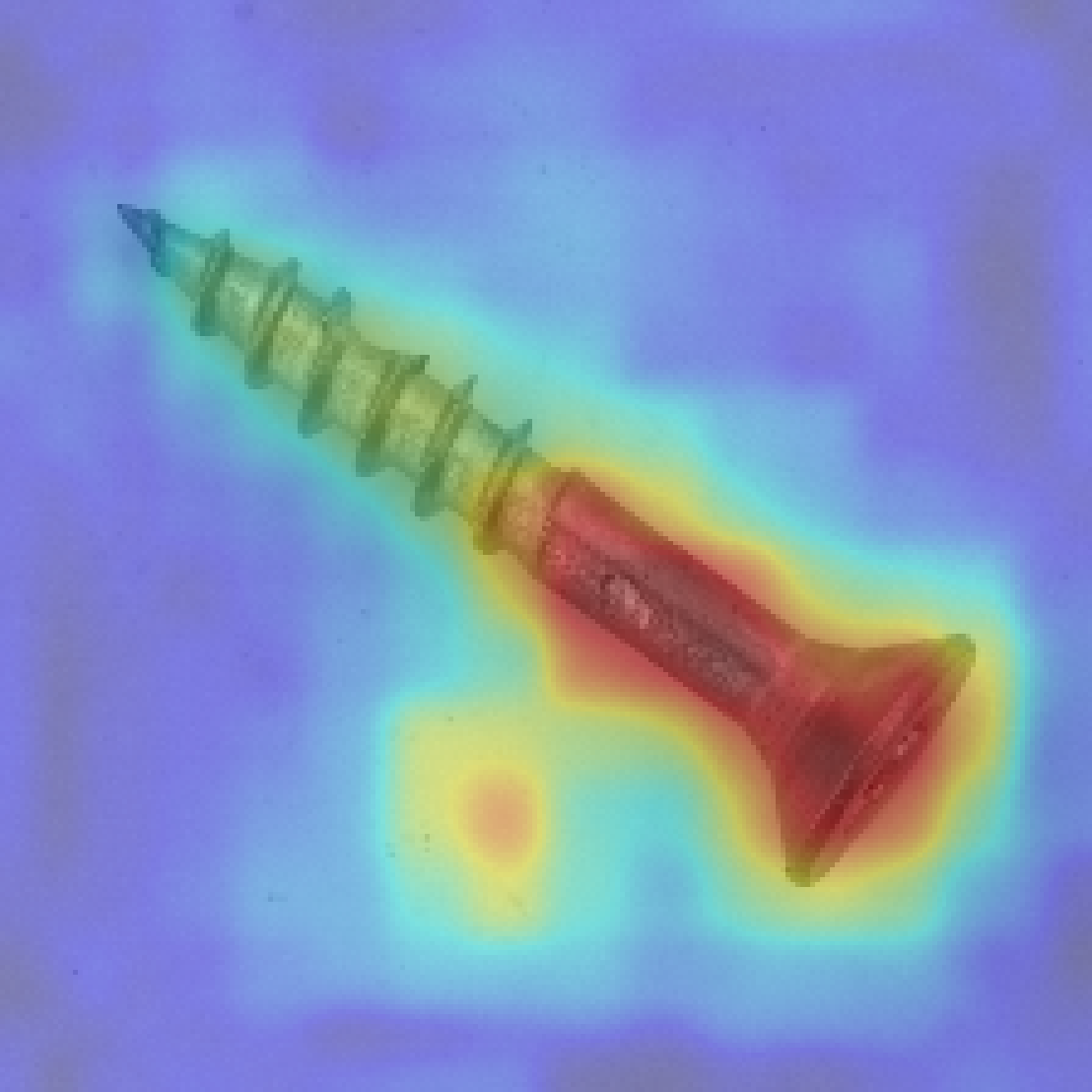}\\
             Input & GT segm. & PatchCore~\cite{roth2022patchcore}  
              & APC (our) \\

\end{tabular}%
\caption{Qualitative localization results for categories \texttt{Pill}, \texttt{Tile}, and \texttt{Screw} on the MVTec dataset for our proposed APC and standard PatchCore with 10 anomalous samples per category. Localization results are overlayed with the original images. Red indicates likely anomalous areas, while blue indicates likely normal areas. \label{fig:picturegrid}}
\end{figure}

On the pixel level, the results in \cref{tab:eval_soa} are different from the image level. PatchCore outperforms APC in AUROC$_{px}$ and F1$_{px}$, while APC leads to a better AUROC$_{px}$ than PRNet. We attribute the loss in localization performance to the emergence of more high-level features, which respond strongly to anomalies. Due to the larger receptive field of these features, segmentations are broader and less defined w.r.t. ground truth regions. The effect is especially evident for the pill example in \cref{fig:picturegrid}. On the other hand, these high-level features positively impact the detection results. Overall, we argue that in industrial settings, anomaly detection is a screening task with subsequent manual intervention. Therefore, a perfect segmentation of the anomaly is usually not necessary and a rough localization is sufficient, which is given by APC as the AUROC$_{px}$ results indicate.

\subsection{Category-level Comparison to State-of-the-art}

\begin{table}[]
\centering
\caption{Per-category results for categories \texttt{Pill}, \texttt{Tile}, and \texttt{Screw} on the MVTec dataset for our proposed APC, standard PatchCore, and PRNet with 10 anomalous samples per category.}
\label{tab:eval_category}
\begin{tabular}{@{}llcccc@{}}
\toprule
Category & System &AUROC$_{im}$ & F1$_{im}$  & AUROC$_{px}$ & F1$_{px}$ \\ \midrule 
\multirow{3}{*}{\texttt{Pill}}&PatchCore~\cite{roth2022patchcore} &  86.46 & 93.18 & 97.27 & 66.30   \\
&PRNet~\cite{zhang2023prototypical} & 83.27 & 90.76 & 96.24 & 66.29  \\
&APC (our) & 97.70 & 96.96 &	95.99 & 47.62  \\ \midrule 
\multirow{3}{*}{\texttt{Tile}}&PatchCore~\cite{roth2022patchcore}  & 100.00 &	100.00 & 95.31 & 60.05   \\
&PRNet~\cite{zhang2023prototypical} & 98.05 &	96.96 &	95.26 &	63.06   \\
&APC (our) & 	100.00 &	100.00 &	97.50 &	69.98 \\  \midrule 
\multirow{3}{*}{\texttt{Screw}}&PatchCore~\cite{roth2022patchcore} &  95.45 & 93.77 & 98.71 & 32.96   \\
&PRNet~\cite{zhang2023prototypical} & 72.14 & 83.63 & 86.22 & 9.33  \\
&APC (our) & 92.59	&91.86&	95.92&	15.55  \\ 
 \bottomrule
\end{tabular}
\end{table}

Further examining the results of APC compared to the state-of-the-art, we present per-category results of three individual object/texture categories of the MVTec dataset in \cref{tab:eval_category}. The first category presented in \cref{tab:eval_category} is \texttt{Pill}, where APC demonstrates the highest improvement over the baseline to 97.70 (+11.27) in AUROC$_{im}$ and to 96.96 (+3.78) in F1$_{im}$ image-level detection scores. As argued before, this comes at the expense of decreased localization performance, due to the stronger and more high-level features. Yet, as visible from the qualitative result in \cref{fig:picturegrid}, the localization is still appropriate to guide manual inspection as typically occurring in the industrial context.

A category where APC improves over the baseline in detection and localization is \texttt{Tile}~(+2.19 AUROC$_{px}$ and +9.92 F1$_{px}$). 
As can be seen from the qualitative sample in \cref{fig:picturegrid}, the improvement mainly comes from decreased false positives. This is similar to other texture categories that uniformly cover the entire image compared to the object categories, where most of the image is covered with background. Therefore,  we assume that the increased amount of relevant original data improves the fine-tuning of the feature extractor in APC.


In contrast to the above-mentioned examples and the general results discussed in \cref{sec:eval_results}, APC does not outperform PatchCore on the category \texttt{Screw} in both detection and localization metrics. This is also visible in the qualitative result in \cref{fig:picturegrid},  where APC is unable to locate the tiny anomalies of the screw. 
We argue that the classification and reconstruction part have difficulties focusing on these tiny image areas leading to sub-optimal results. Note that despite these results, APC still outperforms PRNet in this category and outperforms PatchCore in anomaly detection across all categories. 

\subsection{Ablation Studies}
\label{sec:eval_abl}
In the following, we present four ablations studies on the influence of different aspects of APC. All studies are carried out on the MVTec dataset.

\subsubsection{Influence of Auxiliary Tasks}
To measure the influence of every auxiliary task added for fine-tuning the feature extractor in APC~(classification, segmentation, and reconstruction), we present the results of successively adding the tasks to APC in \cref{tab:eval_abl_tasks}. The results show that adding the classification task improves the image-level detection results in terms of AUROC$_{im}$ over the baseline, which is PatchCore with the MoCo v3 initialization. However, the localization performance~(AUROC$_{px}$ and F1$_{px}$) declines. Adding the segmentation task as the second auxiliary task has only a slightly negative effect on the image-level detection~(AUROC$_{im}$ and F1$_{im}$), but substantially improves the localization performance in terms of AUROC$_{px}$ and F1$_{px}$. Finally, adding the reconstruction task substantially improves the image-level detection at the expense of only a minor loss in localization performance. Overall, each new auxiliary task improves the performance in one aspect, leading to strong overall results in anomaly detection. Yet, no auxiliary task is able to improve both, detection and localization.

\begin{table}[]
\centering
\caption{Results of APC with different combinations of auxiliary tasks during fine-tuning of the feature extractor on the MVTec dataset.} 
\label{tab:eval_abl_tasks}
\begin{tabular}{@{}lcccc@{}}
\toprule
Task(s) in APC &AUROC$_{im}$ & F1$_{im}$  & AUROC$_{px}$ & F1$_{px}$ \\ \midrule
None (baseline) &97.77&	97.13&	97.86&	56.57\\
Class. & 98.05 & 96.62 & 94.07 & 36.68 \\
Class. + Segm. & 97.96 & 96.38 & 95.02 & 40.06\\
Class. + Segm. + Recon. & 98.83 & 97.87 &  94.45 & 39.04  \\ \bottomrule
\end{tabular}
\end{table}


\subsubsection{Architecture for Auxiliary Tasks}
To validate the architecture for the feature extractor's auxiliary tasks in APC, namely the branches for classification, segmentation, and reconstruction, we compare the U-Net-based architecture with two skip-connections~(U-Net$_2$) between the encoder~(feature extractor) and the decoder as described in \cref{sec:method_fe}, with a vanilla U-Net architecture~(U-Net$_4$) featuring four skip connections (dotted lines in Fig.~\ref{fig:sysFig}) to allow more high-level feature to contribute to the auxiliary outputs during the fine-tuning. The results are presented in \cref{tab:eval_abl_arch} and indicate that the additional skip connections between the encoder and the decoder for the auxiliary tasks do not aid the detection results. These results favoring U-Net$_2$ as used in APC match the original PatchCore architecture, which only uses the features from the second and third layers of the feature extractor~(stages 3 and 4 of the ResNet-50). Yet, the localization results improve with more connections, since finer details are more relevant for localization. We still chose the U-Net$_2$ architecture for APC to match PatchCore's architecture and improve the detection performance since it's more relevant in industrial applications~(see \cref{sec:eval_results}).

\begin{table}[]
\centering
\caption{Results of APC with different architectures for the feature extractor during fine-tuning  on the MVTec dataset.}
\label{tab:eval_abl_arch}
\begin{tabular}{@{}lcccc@{}}
\toprule
Architecture of APC &AUROC$_{im}$ & F1$_{im}$  & AUROC$_{px}$ & F1$_{px}$ \\ \midrule
U-Net$_2$ & 98.83 & 97.87 &  94.45 & 39.04 \\ 
U-Net$_4$ & 98.17 &   96.93 &   95.87 &   41.42  \\ \bottomrule 
\end{tabular}
\end{table}

\subsubsection{Pre-training Task for Feature Extractor}
For pre-training the feature extractor, most feature-embedding methods in anomaly detection utilize standard ImageNet dataset with classification task. In contrast, APC utilizes modern self-supervised MoCo v3 pretraining~\cite{mocov3} on ImageNet to improve the feature quality. We compare the two variations of pre-training in APC and report the results in \cref{tab:init}. While ImageNet pretraining has slightly better localization results, the image-level detection results improve with the MoCo v3-based fine-tuning. As discussed above, in industrial applications, detection is more important than detailed localization. Hence, we use MoCo v3 pre-training in APC.

\begin{table}[]
\centering
\caption{Results of APC with different pre-training for the feature extractor during fine-tuning  on the MVTec dataset.}
\label{tab:init}
\begin{tabular}{@{}lcccc@{}}
\toprule
Initialization of APC &AUROC$_{im}$ & F1$_{im}$  & AUROC$_{px}$ & F1$_{px}$ \\ \midrule
ImageNet &   98.00 &   96.84 &   95.98 &   43.60\\
MoCo v3 & 98.83 & 97.87 &  94.45 & 39.04 \\ \bottomrule
\end{tabular}
\end{table}

\subsubsection{Influence of Data Augmentation}

Finally, we investigate the influence of the data augmentation techniques utilized in APC on the results. We suspect that the gap between the reported performance and our reproduction results of PRNet is largely due to differences in synthetic data generation. This leads us to believe that PRNet heavily relies on data augmentation and the generation of synthetic anomalous samples to achieve its reported performance levels. The motivation for this study is to determine whether the same dependence on data augmentation exists for APC. Table~\ref{tab:eval_abl_da} shows the results of APC with only basic augmentations, where no additional anomalous samples are generated, and the results with CutPaste augmentation, where additional anomalous samples are generated by pasting anomalies from an anomalous sample to a normal sample~(see \cref{sec:meth_impl_details}). The results indicate that even without synthetic data generation~(CutPaste), the performance of APC does not drop critically, unlike PRNet, 
therefore proving less dependence on the quality of synthetic data. We hypothesize that this difference comes from the use of the self-supervised task of reconstruction, thus more efficiently utilizing in-domain data. 

\begin{table}[]
\centering
\caption{Results of APC with different data augmentations during fine-tuning  on the MVTec dataset.}
\label{tab:eval_abl_da}
\begin{tabular}{@{}lcccc@{}}
\toprule
Augmentations in APC &AUROC$_{im}$ & F1$_{im}$  & AUROC$_{px}$ & F1$_{px}$ \\ \midrule
Basic & 98.13 & 96.91 & 94.61 & 38.06 \\
Basic + CutPaste & 98.83 & 97.87 &  94.45 & 39.04  \\ \bottomrule
\end{tabular}
\end{table}

\section{Conclusion}
In this paper, we proposed a novel system for industrial anomaly detection, APC, that effectively utilizes knowledge from a few anomalous samples. This is done by integrating a new fine-tuning strategy for the feature extractor based on the three auxiliary tasks classification, segmentation, and reconstruction into a feature-based anomaly detection pipeline. The results of our extensive evaluation on the MVTec dataset show a superior anomaly detection performance compared to state-of-the-art systems, while the localization needs to be further improved, despite already good quality. Yet, this is of secondary interest in industrial anomaly detection due to the screening character with subsequent manual inspection. Further research directions that could address this shortcoming include a better synthetic data generation pipeline.

\subsubsection*{Acknowledgments}
This project was funded by and conducted at Basler AG, Germany, as part of the first author’s master thesis at the University of Hamburg.
\bibliographystyle{splncs04}
\bibliography{egbib}

\end{document}